\pdfoutput=1

\documentclass[11pt]{article}

\usepackage[]{acl}

\usepackage{times}
\usepackage{latexsym}

\usepackage[T1]{fontenc}

\usepackage[utf8]{inputenc}

\usepackage{microtype}

\usepackage{inconsolata}

\usepackage{graphicx}
\usepackage{hyperref}
\usepackage{multirow}
\usepackage{booktabs}
\usepackage{amsmath}
\usepackage{arydshln}
\usepackage{CJKutf8}
\usepackage{graphicx}
\usepackage[normalem]{ulem}
\usepackage{xcolor}
\usepackage{xspace}

\usepackage{pgfplots}
\pgfplotsset{compat=1.8}
\usepgfplotslibrary{statistics}
\usepackage{subfigure}
\usepackage{adjustbox}

\newcommand{\tree}{\textsc{TasTe}\xspace}
\newcommand{\nl}{\textcolor{gray}{\textbackslash n}}

\title{\tree: Teaching Large Language Models to Translate \\ through Self-Reflection}

\author{Yutong Wang$^1$\thanks{~Work was done when Yutong Wang was interning at Pattern Recognition Center, WeChat AI, Tencent Inc, China.}~~
        Jiali Zeng$^2$~~
        Xuebo Liu$^1$\thanks{~Xuebo Liu is the corresponding author.}~~
        Fandong Meng$^2$~~
        \textbf{Jie Zhou}$^2$~~
        \textbf{Min Zhang}$^1$ \\
        $^1$Institute of Computing and Intelligence, Harbin Institute of Technology, Shenzhen, China \\
        $^2$Pattern Recognition Center, WeChat AI, Tencent Inc, China \\ 
        \texttt{wangyutong@stu.hit.edu.cn, \{liuxuebo,zhangmin2021\}@hit.edu.cn} \\
        \texttt{\{lemonzeng,fandongmeng,withtomzhou\}@tencent.com}
}

\begin{document}
\maketitle
\begin{abstract}
Large language models (LLMs) have exhibited remarkable performance in various natural language processing tasks.
Techniques like instruction tuning have effectively enhanced the proficiency of LLMs in the downstream task of machine translation.
However, the existing approaches fail to yield satisfactory translation outputs that match the quality of supervised neural machine translation (NMT) systems.
One plausible explanation for this discrepancy is that the straightforward prompts employed in these methodologies are unable to fully exploit the acquired instruction-following capabilities.
To this end, we propose the \textbf{\tree} framework, which stands for translating through self-reflection.
The self-reflection process includes two stages of inference.
In the first stage, LLMs are instructed to generate preliminary translations and conduct self-assessments on these translations simultaneously.
In the second stage, LLMs are tasked to refine these preliminary translations according to the evaluation results.
The evaluation results in four language directions on the WMT22 benchmark reveal the effectiveness of our approach compared to existing methods.
Our work presents a promising approach to unleash the potential of LLMs and enhance their capabilities in MT.
The codes and datasets are open-sourced at \url{https://github.com/YutongWang1216/ReflectionLLMMT}.
\end{abstract}

\begin{table}[t]
    \centering
    \adjustbox{center}{
        \scalebox{0.85}{
            \begin{tabular}{lp{7.5cm}}
                \toprule
                \textbf{SRC} & \begin{CJK}{UTF8}{gbsn} 但/But\quad 口水鸡/Poached Chicken in Chili Sauce\quad 整盒/whole box\quad 是/is\quad 生肉/row meat，没办法/cannot\quad 吃/eat，是否/whether\quad 可以/can\quad 退款/refund? \end{CJK} \\
                \textbf{REF} & But the Poached Chicken in Chili Sauce is just a box of inedible raw chicken, and can I get a refund for it? \\
                \hdashline
                \textbf{Normal} & But the whole box of chicken is raw meat, so you can't eat it, can you refund? \\
                \hdashline
                \textbf{Stage 1} & But \textcolor{red}{\sout{the whole box is meat}}, can't eat, can you refund \textbackslash n\colorbox{yellow}{[Bad]} \\
                \textbf{Stage 2} & But the whole box of spicy chicken is raw meat, so I can't eat it. Can I get a refund? \\
                \bottomrule
            \end{tabular}
        }
    }
    \caption{An example of the \tree approach. ``Normal'' denotes the output of the baseline LLM fine-tuned on a normal parallel corpus. ``Stage 1'' and ``Stage 2'' denote the outputs of the first and second inference stages of the proposed self-reflection process, respectively. The \colorbox{yellow}{highlight} denotes the quality label predicted by the LLM. Inherent translation errors generated in the first stage, such as the \textcolor{red}{\sout{red strikethrough}} part, are rectified in the second inference stage.}
    \label{tab:reflect}
\end{table}

\section{Introduction}
Large language models (LLMs) like GPT-4 \citep{openai2023gpt4} have recently demonstrated dramatic performance across a wide range of natural language processing tasks \citep{bubeck2023sparks, liang2022holistic}.
Their outstanding grasp of syntactic and semantic knowledge positions them as powerful instruments for the enhancement of machine translation, capable of producing translations of superior quality \citep{hendy2023good, zhang2023prompting, garcia2022using}.
This substantial progress represents an evolution of the paradigm in machine translation, serving as the foundation of novel translation systems characterized by enhanced quality and reliability.

Numerous studies are underway to unlock the vast potential of machine translation within LLMs.
Prompt engineering aims to design effective prompt templates to guide LLMs in accomplishing specific language tasks.
Some approaches attempt to integrate additional information relevant to the translation task to enhance the performance of LLMs \citep{ghazvininejad2023dictionary, lu2023chain, he2024exploring, peng-etal-2023-towards}.
Studies in In-Context Learning (ICL, \citealp{brown2020language}) seek to provide LLMs with more relevant and high-quality translation exemplars, which assists LLMs in retrieving bilingual knowledge, facilitating the generation of translations of the highest possible quality \citep{vilar2023prompting, agrawal2023context}.
However, assessments of LLMs reveal that, in most translation directions, their performance falls short of that exhibited by robust supervised baselines \citep{zhu2023multilingual}.
This shortfall is due to the fact that these approaches often treat the LLM machine translation task as a simple text generation task, focusing on adjusting the prompts to enhance the outcomes.
However, the intrinsic features of the machine translation task, such as the need for diverse multilingual knowledge, are often overlooked.

Some studies recommend the tuning of relatively smaller LLMs for translation \citep{zhu2023multilingual, xu2023paradigm}.
Instruction tuning of LLMs with a limited number of high-quality supervised instructions in machine translation tasks yields remarkable results in some instances \citep{zeng2023tim, jiao2023parrot, zhu2023multilingual, hendy2023good}.
Despite these achievements, these attempts still fail to fully leverage the capacity of LLMs due to their overly straightforward inference process.
Unlike supervised NMT models, LLMs generate translations through language modeling, which contains a more complicated inference process and relies more on inherent linguistic knowledge.
Studies such as Chain-of-Thought (CoT) reveal that the introduction of intermediate reasoning steps in the inference process significantly increases the reasoning capabilities of language models \citep{wei2022chain, kojima2022large}.

In this paper, we introduce \textbf{\tree}, a method that aims at improving the translation performance of LLMs by instilling the ability to self-reflect on their own outputs.
Specifically, we segment the LLM translation process into two stages of inference.
In the first stage, LLMs are prompted to generate preliminary translations while simultaneously making quality predictions for these translations.
In the second stage, we instruct LLMs to refine these preliminary translations based on the predicted quality levels to produce final candidates.
An example of the proposed process can be found in Table \ref{tab:reflect}.
This entire process can be regarded as a form of self-reflection, mirroring the common approach employed by humans to carry out tasks more effectively and impeccably.
To establish a sufficient multitask capability for executing the entire reflective translation process, we conduct supervised fine-tuning (SFT) on LLMs using a multi-task training dataset.
This method demonstrates a remarkable stimulation of the potential of LLMs, providing a novel approach to enhance the translation performance of these models.

Our contributions are summarized as follows:
\begin{itemize}
    \item We present the \textbf{\tree} method, which guides LLMs through a two-stage inference process, allowing them to initially generate preliminary results and subsequently refine them into improved candidates based on their self-assessment results.
    \item We create a multi-task training set comprising tasks that are closely aligned with the \tree process to equip LLMs with the capability to execute the whole inference process.
    \item 
    We find that by employing the \tree method, LLMs proficiently refine their initial translation candidates, resulting in superior final outcomes, which in turn contributes to an enhancement in their translation capabilities.
\end{itemize}

\begin{figure*}[t]
    \centering
    \includegraphics[width=1.0\linewidth]{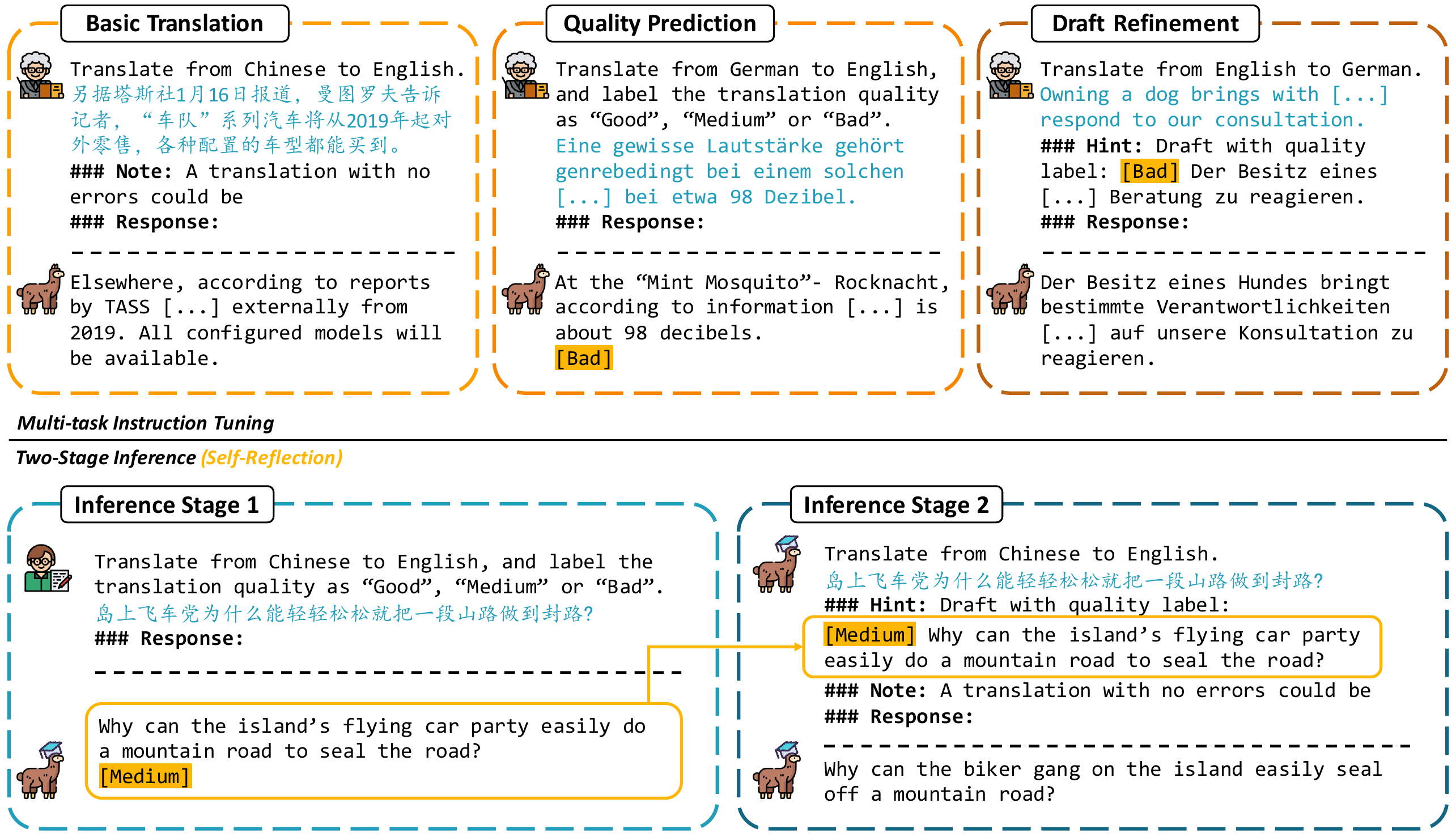}
    \caption{The framework of our proposed \tree method.}
    \label{fig:framework}
\end{figure*}

\section{Related Work}
Efforts to enhance the translation performance of LLMs can be categorized into two research lines: prompt engineering and instruction tuning.
Prompt Engineering aims to design proper prompt templates and introduce prior knowledge or supplementary information to support the inference process.
Dictionary-based approaches incorporate control hints in the prompt from bilingual or multilingual dictionaries to deal with rare words in source sentences \citep{ghazvininejad2023dictionary, lu2023chain}.
\citet{he2024exploring} extracts translation-related knowledge, such as topics, by self-prompting to guide the translation process.
Studies in ICL (\citealp{brown2020language}) aim to provide LLMs with more relevant and high-quality translation exemplars.
This approach assists LLMs in retrieving bilingual knowledge, facilitating the generation of translations of the highest possible quality \citep{vilar2023prompting, agrawal2023context}.

Instruction tuning represents an efficient method to enhance the ability of LLMs to follow natural language instructions and yield outputs that align more closely with human preference in downstream zero-shot tasks \citep{wei2021finetuned, NEURIPS2022_b1efde53, chung2024scaling}.
\citet{jiao2023parrot} explore several translation instructions to improve the translation performance of LLMs.
\citet{zeng2023tim} employ examples in comparison to instruct LLMs and calculate the additional loss.
\citet{zhang2023bayling} enhance the multilingual language generation and instruction following capabilities of LLMs through interactive translation tasks.

Additionally, several studies proposed to facilitate a similar reflection process, utilizing confidence-guided approaches or multi-step inference, to assist the translation procedure.
\citet{lu2022learning} train a confidence estimation network in parallel with the backbone network to predict the confidence levels for generated translations, determining the amount of hints the model requires to produce correct translations.
\citet{xia2017deliberation} introduce a second-pass decoder to the conventional encoder-decoder structure, polishing the initial drafts and generating the final outputs.
\citet{tan2022msp} divide the translation process into three stages and independently apply different continuous prompts to better shift language to translation tasks.
\citet{li2023deliberate} propose a deliberate-then-generate inference framework, where LLMs are first prompted to detect error types from given candidates and then generate their final answers.
\citet{chen2023iterative} propose to iteratively prompt LLMs to self-correct their translations.
\citet{feng2024improving} introduce a self-correcting inference framework for LLMs accessible via APIs, where LLMs autonomously conduct MQM self-evaluations and refine the primary candidates based on the evaluation results.
\citet{ki2024guiding} utilize a trained fine-grained feedback model to identify defects in generated translations, subsequently directing LLMs to refine the translations based on the feedback.

Our work represents a fusion of instruction tuning and the CoT methodology.
We introduce a multi-step inference translation process in imitation of the self-reflection mechanism observed in humans.
The utilization of multitask training data, including Basic Translation, Quality Prediction, and Draft Refinement, substantiates not only the multi-step inference capability but also the comprehension of nuances in translation quality.

\section{\tree: Translate through Reflection}
\subsection{Overall Framework}
In this work, we aim to enhance the translation capabilities of LLMs by instructing them to engage in self-reflection on their translation candidates, ultimately producing carefully refined outputs.
This process is achieved through a two-stage inference.

In the first stage, we ask the models to generate preliminary translations.
Different from the conventional machine translation process, we also require them to predict the quality of their own outputs simultaneously.
These preliminary translations are named ``drafts'', and their corresponding quality predictions can take the form of either approximate labels or precise scores.
This stage of inference can be formalized into the following formula:
\begin{equation}
    (\boldsymbol{y},q) \sim P(\boldsymbol{y},q \mid \boldsymbol{w},\boldsymbol{x};\theta)
\end{equation}
\begin{equation}
\begin{split}
    & P(\boldsymbol{y}_{1:m},q \mid \boldsymbol{w},\boldsymbol{x};\theta) \\
    =& P(q \mid \boldsymbol{y}_{1:m},\boldsymbol{w},\boldsymbol{x};\theta)P(\boldsymbol{y}_{1:m} \mid \boldsymbol{w},\boldsymbol{x};\theta) \\
    =& P(q \mid \boldsymbol{y}_{1:m},\boldsymbol{w},\boldsymbol{x};\theta)\prod_{t=1}^m{P(\boldsymbol{y}_i \mid \boldsymbol{y}_{1:t-1},\boldsymbol{w},\boldsymbol{x};\theta)}
\end{split}
\end{equation}
where $\theta$ represents the parameters of the LLM, $\boldsymbol{x}$ and $\boldsymbol{w}$ denote the source sentence and the rest of the prompt (including the instruction), respectively.
The preliminary translation $\boldsymbol{y}_{1:m}$ is generated first, and the quality label (score) $q$ is generated later according to $\boldsymbol{y}_{1:m}$.
The corresponding prompts of the first inference stage are illustrated in the ``Inference Stage 1'' box in Figure \ref{fig:framework}.

In the second stage, we guide the models to refine their drafts based on the quality predictions.
Both the drafts and quality labels/scores are formatted into the input field of the prompts for LLMs.
The models proceed to make appropriate adjustments to the drafts according to the predicted label/scores, yielding the final translation candidates in a refined form.
This stage of inference can be formalized into the following formula:
\begin{equation}
    \boldsymbol{y}' \sim P(\boldsymbol{y}' \mid \boldsymbol{y},q,\boldsymbol{w}',\boldsymbol{x};\theta)
\end{equation}
\begin{equation}
\begin{split}
    & P(\boldsymbol{y}'_{1:n} \mid \boldsymbol{y},q,\boldsymbol{w}',\boldsymbol{x};\theta) \\
    =& \prod_{t=1}^n{P(\boldsymbol{y}'_i \mid \boldsymbol{y}'_{1:t-1},\boldsymbol{y},q,\boldsymbol{w}',\boldsymbol{x};\theta)}
\end{split}
\end{equation}
where $\boldsymbol{w}'$ denotes the new prompt employed in the second stage.
The refined translation $\boldsymbol{y}'_{1:n}$ is generated according to the preliminary translation $\boldsymbol{y}$ with its predicted quality level $q$.
The corresponding prompts of the second inference stage are shown in the ``Inference Stage 2'' box in Figure \ref{fig:framework}.

\subsection{Multitask SFT}\label{subsec:multitask}
To ensure that LLMs achieve a comprehensive understanding of the task instructions, we conduct multitask SFT on the models.
The multitasking approach consists of three components: \textbf{Quality Prediction}, \textbf{Basic Translation}, and \textbf{Draft Refinement}.

\paragraph{Quality Prediction}
In this sub-task, LLMs are tasked with generating translations and providing self-quality predictions for a given source sentence.
The quality prediction task consists of two forms: a) Text Classification (TC), entailing label predictions of ``Good'', ``Medium'', or ``Bad'', and b) Quality Estimation (QE), involving integer score prediction ranging from 0 to 100.
We utilize candidates of various qualities generated by multiple systems, along with their evaluated COMET scores, to construct fine-tuning instances.
Please refer to Appendix \ref{sec:quality_prediction} for detailed information.
The ground truth of the training data would be translations with gold quality labels/scores placed in the back.

\paragraph{Basic Translation}
We utilize parallel data combined with a standardized instruction to conduct fine-tuning of LLMs for multilingual translation tasks, including German $\Leftrightarrow$ English and Chinese $\Leftrightarrow$ English language pairs .
The instruction is formulated straightforwardly as ``\texttt{Translate from [SRC] to [TGT]}''.
As shown in Figure \ref{fig:framework}, the Basic Translation instructions exhibit a high degree of similarity to their Quality Prediction counterparts, but they belong to two completely different tasks.
To disambiguate instructions between these two tasks and prevent LLMs from obtaining low-quality translation knowledge, we follow \citet{zeng2023tim} to append a distinguishing note ``\texttt{\#\#\# Note: A translation with no errors could be}'' at the end of the Basic Translation input.

\paragraph{Draft Refinement}
In this sub-task, LLMs are asked to refine drafts based on quality labels/scores to produce final outputs.
Given a source sentence and multiple candidates of various qualities, we designate the highest-scored output as the reference.
The drafts are sampled from the remaining candidates, covering all quality levels.
We incorporate a new field named ``\texttt{Hint}'' within the translation prompt.
This field provides LLMs with translation drafts of the source sentence, with quality labels/scores placed in front of the drafts in the following format: ``\texttt{\#\#\# Hint: Draft with quality label/score: [LABEL/SCORE] [Draft]}''.
We fill in ``label'' or ``score'' based on whether the TC or QE approach is employed.
Examples of the complete prompts are shown in Table \ref{tab:prompt_example}.

\begin{table*}[t]
    \centering
    \scalebox{0.8}{
    \begin{tabular}{lcccccccccc}
        \toprule
        \multirow{2}{*}{\textbf{System}} & \multicolumn{2}{c}{\textbf{Zh$\Rightarrow$En}} & \multicolumn{2}{c}{\textbf{En$\Rightarrow$Zh}} & \multicolumn{2}{c}{\textbf{De$\Rightarrow$En}} & \multicolumn{2}{c}{\textbf{En$\Rightarrow$De}} & \multicolumn{2}{c}{\textbf{Average}} \\
        \cmidrule(lr){2-3} \cmidrule(lr){4-5} \cmidrule(lr){6-7} \cmidrule(lr){8-9} \cmidrule(lr){10-11}
         & COMET & BLEU & COMET & BLEU & COMET & BLEU & COMET & BLEU & COMET & BLEU \\
        \midrule
        WMT22 Winners & 81.00 & 33.50 & 86.80 & 54.30 & 85.00 & 33.70 & 87.40 & 38.40 & 85.05 & 39.98 \\
        NLLB-3.3b & 76.92 & 21.07 & 81.56 & 32.52 & 83.42 & 29.54 & 86.23 & 33.98 & 82.03 & 29.28 \\
        \midrule
        \multicolumn{11}{c}{Backbone: \texttt{LLaMA}} \\
        ParroT & 75.90 & 20.20 & 80.30 & 30.30 & 82.40 & 27.30 & 81.60 & 26.10 & 80.05 & 25.98 \\
        Bayling & 77.48 & 20.31 & \textbf{84.43} & \textbf{38.19} & 83.19 & 28.16 & 82.18 & 25.66 & 81.82 & 28.08 \\
        MT-\textit{Full} & 78.72 & 23.80 & 83.35 & 33.01 & 83.79 & 30.10 & 83.70 & 27.18 & 82.39 & 28.52 \\
        MT-\textit{FixEmb} & 79.02 & 24.30 & 83.62 & 33.33 & 84.05 & 30.62 & 83.66 & 27.75 & 82.59 & 29.00 \\
        \hdashline
        \tree & & & & & & & & \\
        \quad\textit{Full-QE} & 79.17 & 24.27 & 83.90 & 34.25 & 83.83 & 30.49 & 83.38 & 27.16 & 82.57 & 29.04 \\
        \quad\textit{Full-TC} & 79.31 & 24.23 & 84.00 & 34.51 & 83.92 & 30.17 & 82.95 & 26.74 & 82.55 & 28.91 \\
        \quad\textit{FixEmb-QE} & 79.35 & 24.47 & 84.30 & 34.94 & 84.07 & 30.75 & 83.70 & 27.32 & 82.86 & 29.37 \\
        \quad\textit{FixEmb-TC} & \textbf{79.53} & \textbf{24.87} & 84.24 & 34.96 & \textbf{84.11} & \textbf{31.03} & \textbf{83.80} & \textbf{27.94} & \textbf{82.92} & \textbf{29.70} \\
        \midrule
        \multicolumn{11}{c}{Backbone: \texttt{BLOOM}} \\
        ParroT & 79.00 & 22.70 & 83.50 & 34.50 & 78.00 & 24.90 & 73.60 & 20.50 & 78.53 & 25.65 \\
        TIM & 79.71 & \textbf{24.51} & 85.10 & 37.83 & 78.94 & 26.12 & 74.91 & 20.90 & 79.67 & 27.34 \\
        MT-\textit{Full} & 79.25 & 22.81 & 85.01 & 35.49 & 77.61 & 24.05 & 71.31 & 18.84 & 78.30 & 25.30 \\
        MT-\textit{FixEmb} & 79.84 & 23.43 & 85.20 & 36.68 & 78.27 & 25.07 & 72.06 & 19.41 & 78.84 & 26.15 \\
        \hdashline
        \tree & & & & & & & & \\
        \quad\textit{Full-QE} & 79.36 & 23.15 & 85.05 & 36.84 & 78.42 & 24.87 & 75.41 & 21.18 & 79.56 & 26.51 \\
        \quad\textit{Full-TC} & 79.14 & 23.04 & 84.94 & 36.75 & 78.74 & 24.97 & 75.53 & 21.13 & 79.59 & 26.47 \\
        \quad\textit{FixEmb-QE} & \textbf{80.40} & 24.41 & 85.81 & \textbf{39.31} & \textbf{79.20} & \textbf{26.28} & 76.30 & 21.84 & \textbf{80.43} & \textbf{27.96} \\
        \quad\textit{FixEmb-TC} & 80.28 & 24.20 & \textbf{85.90} & 39.07 & 78.96 & 26.27 & \textbf{76.38} & \textbf{21.98} & 80.38 & 27.88 \\
        \bottomrule
    \end{tabular}}
    \caption{Main results of \tree. \texttt{LLaMA-2-7b} and \texttt{BLOOMZ-7b1-mt} are chosen as the backbone model. \textit{QE} and \textit{TC} signify that the Quality Prediction subtask takes the form of quality estimation and text classification, respectively. The best results of each kind of backbone model are labeled using \textbf{bold font}.}
    \label{tab:main_results}
\end{table*}

\section{Experimental Setups}

\subsection{Data}
We employ the WMT validation set to construct the training data for the Basic Translation task and utilize the MTME multi-candidate\footnote{\href{https://github.com/google-research/mt-metrics-eval}{https://github.com/google-research/mt-metrics-eval}} dataset, which contains source sentences and their candidate translations generated by multiple systems to build the training data for the Quality Prediction and Draft Refinement tasks.
For Quality Prediction, candidates across various quality levels are sampled to form training instances.
For Draft Refinements, the candidate with the highest COMET score is chosen as the reference, and the drafts to be refined are sampled from the other candidates covering various qualities.
The data statistics and details of data building can be found in Appendix \ref{sec:data_details}.

To avoid possible data leakage in the training data, we evaluate the translation performance on the WMT22 test set \citep{kocmi2022findings}, which covers domains such as news, social, e-commerce, and conversation.
We present the translation results in German $\Leftrightarrow$ English and Chinese $\Leftrightarrow$ English directions.
We report the BLEU scores by SacreBLEU ~\citep{post-2018-call} and COMET scores by \texttt{wmt22-comet-da} \citep{rei-etal-2022-comet}.

\subsection{Model Training}
We employ \texttt{BLOOMZ-7b1-mt}\footnote{\href{https://huggingface.co/bigscience/bloomz-7b1-mt}{https://huggingface.co/bigscience/bloomz-7b1-mt}} and \texttt{LLaMA-2-7b}\footnote{\href{https://huggingface.co/meta-llama/Llama-2-7b}{https://huggingface.co/meta-llama/Llama-2-7b}} \citep{touvron2023llama} as our backbone models.
These models are all fine-tuned for 1 epoch with a batch size of 128.
The learning rates are set to 2e-5, and the weight decay parameter is set to 0.0.
The maximum text length is 768. We conducted the fine-tuning on eight NVIDIA A100 GPUs, using the Deep-Speed ZeRO stage3 for acceleration.

We employ two distinct training strategies, differing in the updated parameters:

\paragraph{Full-Parameter Tuning (Full)}
In this method, all the parameters in LLMs are involved in the training process.
In comparison to methods that focus on training only a small set of parameters (such as Prefix Tuning and Low-Rank Adaption), full-parameter tuning is less susceptible to overfitting due to the larger parameter space.
However, the main issue with this approach is excessive memory consumption and runtime demands.

\paragraph{Tuning with Fixed Embedding Layer (FixEmb)}
The embedding layer is pre-trained on large-scale corpus and reflects the general distribution of word embeddings.
Further tuning, especially when the number of trainable parameters is limited or the training corpus is not abundant enough, will introduce disturbances into these distributions, leading to a decline in the model's expressive capacity.
To overcome this problem, we freeze the embedding layers of LLMs and fine-tune the rest of the parameters.
This assists LLMs in maintaining correctness and diversity in their expressions.

\subsection{Baselines}
The \textbf{MT-($\cdot$)} baseline models represent the LLMs trained exclusively with the Basic Translation dataset, as outlined in Table \ref{tab:data_size}.
This dataset contains the German $\Leftrightarrow$ English and Chinese $\Leftrightarrow$ English translation directions.

Additionally, we present the results of WMT22 winners, \texttt{NLLB-3.3B} \citep{costa2022no}, a multilingual translation model trained in over 200 languages,
\texttt{Bayling} \citep{zhang2023bayling}, \texttt{ParroT} \citep{jiao2023parrot}, and \texttt{TIM} \citep{zeng2023tim}, LLMs fine-tuned for machine translation with \texttt{BLOOM} or \texttt{LLaMA} as the backbone models.

\section{Results} \label{sec:results}
Our main results are shown in Table \ref{tab:main_results}.
Almost all of our methods outperform the corresponding MT-($\cdot$) baseline across both metrics and all language pairs, providing evidence of the effectiveness of our approach in enhancing the translation capabilities of LLMs.
When utilizing \texttt{BLOOMZ-7b1-mt} as the backbone model, our \textit{FixEmb}-($\cdot$) approaches achieve favorable results, particularly in Zh $\Leftrightarrow$ En directions, and outperform \texttt{ParroT} and \texttt{TIM} across all language pairs on COMET scores.
While employing \texttt{LLaMA-2-7b} as the backbone model, our \textit{FixEmb}-($\cdot$) approaches also gain remarkable results, particularly in De $\Leftrightarrow$ En directions, and beat \texttt{Bayling} in all directions except En $\Leftrightarrow$ Zh.

There is no significant difference in translation performance observed between two different quality prediction approaches, ($\cdot$)-\textit{QE} and ($\cdot$)-\textit{TC}.
This suggests that both of these approaches effectively aid LLMs in grasping the quality differences between varying translations.

The models trained with fixed embedding layers consistently outperform their counterparts trained with full parameters across all language pairs and both evaluation metrics.
We argue that this is because fixing embedding layers during fine-tuning effectively preserves the expressive capability of LLMs against word distribution biases within the training data.
This facilitates the generalization of LLMs across the word domain, mitigating over-fitting and thereby enhancing their capacity to produce robust and diverse translations.

We also train a merged model that handles QE and TC approaches simultaneously, and conduct a comparison of the translation performance across models of different scales.
Please refer to Appendix \ref{sec:merged_model} and \ref{sec:model_scaling} for more details.

\section{Analysis}
Unless mentioned otherwise, the subsequent experiments are conducted in the \textit{FixEmb-TC} setting.

\begin{table}[t]
    \centering
    \small
    \begin{tabular}{lccccc}
        \toprule
        Model & PPL & Pred.$\uparrow$ & P$\uparrow$ & R$\uparrow$ & F1$\uparrow$ \\
        \midrule
        BLOOMZ & -37.10 & 76.84 & 70.1 & 68.2 & 67.6 \\
        LLaMA-2 & 0.00 & 80.33 & 70.5 & 70.1 & 69.8 \\
        \bottomrule
    \end{tabular}
    \caption{Evaluation results on quality prediction task in Zh $\Rightarrow$ En direction. Precision, recall, and F1 values are calculated as weighted averages across three translation quality categories. PPL/Pred. represents Pearson's $r$ between the perplexity values/predicted scores and the COMET scores.}
    \label{tab:howgood}
\end{table}

\subsection{How Good Are LLMs at Quality Prediction?}
Quality Prediction constitutes an end-to-end process, where LLMs are instructed to predict quality labels or scores while generating translations.
To validate the assertion that LLMs have genuinely acquired the capability to predict the quality of candidates, we evaluated the quality prediction outputs.
For TC, we construct gold labels for the instances according to their COMET scores following the same principle mentioned in Appendix \ref{sec:data_details} and report the precision, recall, and F1 values of the predicted labels.
For QE, we assessed the Pearson's correlation coefficient between the predicted quality scores and the gold COMET scores.
Additionally, we present the Pearson's correlation coefficient between the perplexity values (PPL) of the candidates and the COMET scores for comparison.

As shown in Table \ref{tab:howgood}, for the TC approach, the models exhibit a commendable level of accuracy in assigning quality labels to their translations, as evidenced by F1 values surpassing 67.6.
In the QE task, our models produce scores with a satisfactory correlation with COMET scores (the p-values are all smaller than 0.01), while the perplexity values demonstrate a relatively poor correlation with COMET scores.
These statistics demonstrate that our models can make precise quality predictions for their own generated translations, providing a dependable reference for the Draft Refinement task.

We can also discover that \texttt{LLaMA-2} outperforms \texttt{BLOOMZ} in terms of accuracy for both the QE and TC tasks, suggesting that \texttt{LLaMA-2} possesses a more extensive bilingual knowledge base.

\subsection{Effect of Draft Refinement}
To analyze the influence of the Draft Refinement process (i.e., the second stage of inference), we perform the following two comparisons between the candidates obtained after the first and second inference stages.

\begin{figure}[t]
    \centering
    \includegraphics[width=0.6\linewidth]{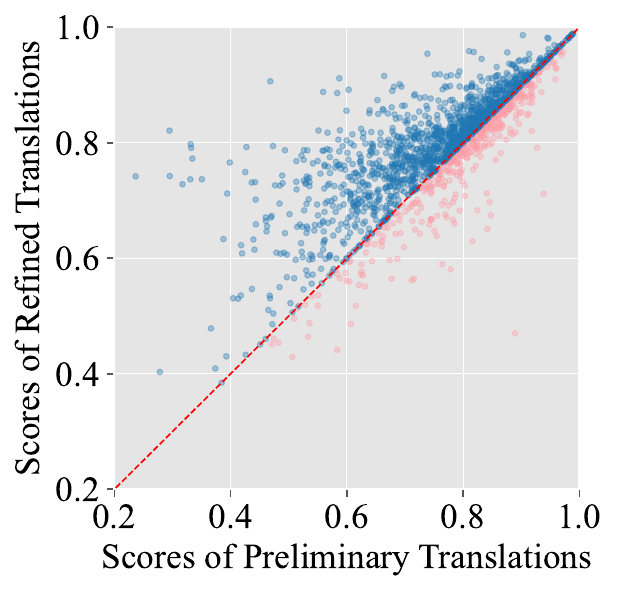}
    \caption{Comparison between the COMET scores of the preliminary and refined translations. The results are obtained by \texttt{LLaMA-2-7b} in Zh$\Rightarrow$En direction.}
    \label{fig:refine}
\end{figure}

\paragraph{Translation Quality}
We evaluate the COMET scores of the preliminary and refined translations.
The results are shown in Figure \ref{fig:refine}.
In the plot, each point located above the diagonal line represents an instance where a quality improvement is achieved through refinement.
As the plot demonstrates, a majority of the final candidates exhibit higher quality levels than their initial counterparts.

\begin{table}[t]
    \centering
    \small
    \begin{tabular}{lcc}
        \toprule
        Label & Proportion (\%) & $\Delta$COMET \\
        \midrule
        Good & 31.89 & 0.45 \\
        Medium & 32.80 & 2.06 \\
        Bad & 35.31 & 7.79 \\
        \bottomrule
    \end{tabular}
    \caption{Proportions of preliminary translations with different predicted quality labels and their average COMET scores increments during refinement. These results are obtained by \texttt{LLaMA-2-7b} in Zh $\Rightarrow$ En direction.}
    \label{tab:score_increment}
\end{table}

\begin{figure}[t]
    \centering
    \includegraphics[width=1.0\linewidth]{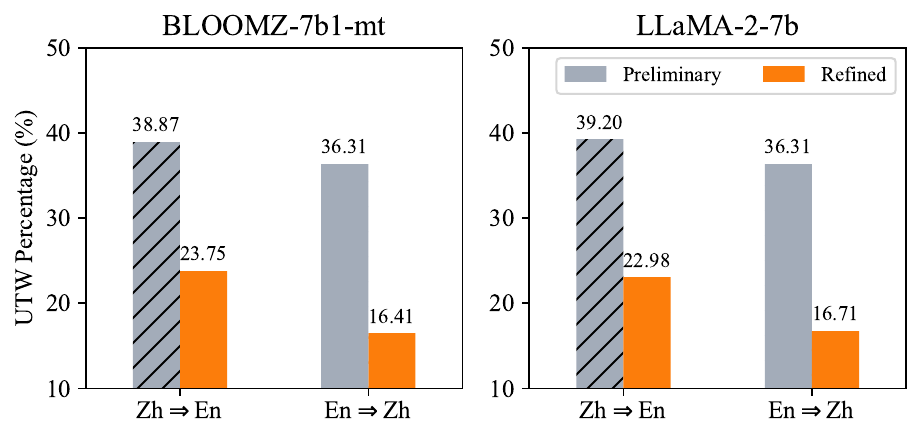}
    \caption{Comparison between the UTW percentages of the preliminary and refined translations.}
    \label{fig:refine_utw}
\end{figure}

Table \ref{tab:score_increment} illustrates the proportions of preliminary translations with varying predicted quality labels and their respective average COMET score increments during the refinement process.
The most significant score enhancements are observed in instances labeled as ``Bad'', which constitute the largest proportion of all instances.
Subsequently, ``Medium'' instances show a moderate improvement,
while ``Good'' instances exhibit the least noticeable enhancement.
These observations highlight the efficacy of the Draft Refinement process in refining the preliminary translations generated in the first inference stage as well as rectifying potential generation failures, as evidenced by instances located in the top-left region of Figure \ref{fig:refine}.

\paragraph{Unaligned Translation Words (UTW)}
We measure the percentages of target words that remain unaligned in a word-to-word alignment between the source sentences and translations obtained after the first and second inference stages.
The alignments are extracted using the tool developed by \citet{dou2021word}.
This measurement is also used by \citet{hendy2023good} to investigate the presence of words that have no support in the source sentences.
The results are shown in Figure \ref{fig:refine_utw}.
We can observe that the amount of UTW is significantly reduced during the draft refinement process, with a decrease of more than 15 percentage points.
This observation suggests that the Draft Refinement process contributes to a reduction in hallucinations within the candidates, leading to a higher level of translation precision and mitigation of potential risks within the translation systems.

\begin{table}[t]
    \centering
    \small
    \begin{tabular}{lcc}
        \toprule
        Label & Edit Distance & COMET \\
        \midrule
        Origin & $\text{18.98}_\text{+0.00}$ & $\text{79.53}_\text{+0.00}$ \\
        Good & $\text{16.95}_\text{-2.03}$ & $\text{79.25}_\text{-0.28}$ \\
        Random & $\text{18.78}_\text{-0.20}$ & $\text{79.36}_\text{-0.17}$ \\
        Bad & $\text{20.20}_\text{+1.22}$ & $\text{79.51}_\text{-0.02}$ \\
        Blank & $\text{18.12}_\text{-0.86}$ & $\text{79.08}_\text{-0.45}$ \\
        \bottomrule
    \end{tabular}
    \caption{The edit distance between the preliminary and refined translations and the final COMET scores under different quality label configurations. ``Origin'' represents the configuration where predicted labels remain unmodified. ``Blank'' represents that quality labels are removed during refinement processes. These results are obtained by \texttt{LLaMA-2-7b} in Zh $\Rightarrow$ En direction.}
    \label{tab:label_role}
\end{table}

\subsection{The Role of Quality Labels}
To examine the impact of the predicted quality labels on the refinement process, we conduct experiments by modifying these labels with the following configurations:
a) All the labels are set to ``Good''.
b) All the labels are set to ``Bad''.
c) All the labels are randomly sampled among ``Good'', ``Medium'', and ``Bad''.
d) All the labels are removed from the prompts and the model is only provided with draft translations during the refinement process.
Subsequently, we perform the refinement process, and calculate the average edit distances between the preliminary and refined translations as follows:
\begin{equation}
    \begin{split}
        \overline{d} &= \frac{1}{n}\sum_{i=1}^{n}{(1 - \text{LevRatio}_i)} \\
        &= \frac{1}{n}\sum_{i=1}^{n}{\frac{\text{LevDist}_i}{\text{len}_{i}^1 + \text{len}_{i}^2}}
    \end{split}
\end{equation}
Here, $\text{LevRatio}_i$ represents the Levenshtein distance radio\footnote{{\href{https://rapidfuzz.github.io/Levenshtein/levenshtein.html\#ratio}{https://rapidfuzz.github.io/Levenshtein/levenshtein.html}}} of the $i$-th instance, $\text{len}_{i}^1$ and $\text{len}_{i}^2$ represent the lengths of two strings, respectively, and $\text{LevDist}_i$ represents the Levenshtein distance between these strings.

\begin{table}[t]
    \centering
    \small
    \begin{tabular}{lcc}
        \toprule
        Method & BLEU & COMET \\
        \midrule
        MT-\textit{FixEmb} & 19.41 & 72.06 \\
        \tree & 21.98 & 76.38 \\
        \quad \textit{w/o Basic Translation} & 20.00 & 72.12 \\
        \quad \textit{w/o Quality Prediction} & 17.86 & 72.26 \\
        \quad \textit{w/o Draft Refinement} & 19.31 & 72.00 \\
        \bottomrule
    \end{tabular}
    \caption{Ablation Study. We report the BLEU and COMET scores in En$\Rightarrow$De direction achieved by \texttt{BLOOMZ-7b1-mt}.}
    \label{tab:ablation}
\end{table}

\begin{table}[t]
    \centering
    \small
    \begin{tabular}{lcccc}
        \toprule
        System & Zh$\Rightarrow$En & En$\Rightarrow$Zh & De$\Rightarrow$En & En$\Rightarrow$De \\
        \midrule
        CoT-7b & 74.50 & 73.79 & 79.63 & 74.37 \\
        CoT-13b & 75.21 & 75.32 & 80.10 & 73.55 \\
        \tree & 79.53 & 84.24 & 84.11 & 83.80 \\
        \bottomrule
    \end{tabular}
    \caption{COMET scores gained by our approach and the CoT method.}
    \label{tab:taste_vs_cot}
\end{table}

We report the average edit distances and the COMET score of the refined translations in Table \ref{tab:label_role}.
In the cases where all the labels are set to ``Good'', the edit distances between the preliminary and refined translations are relatively small.
This suggests that the model tends to make fewer modifications to the preliminary translations.
Conversely, when all the labels are set to ``Bad'', the edit distances are relatively large, indicating that the model tends to make more modifications during refinement.
Furthermore, noticeable performance decreases are observed when the labels are set to ``Good'', sampled randomly (i.e. Random), or removed from the prompts (i.e. Blank).
These phenomena illustrate the impact of the quality labels in the refinement process, which is to assist LLMs in making reasonable adjustments based on the actual translation quality levels and generating high-quality final candidates.

\subsection{Ablation Study}
To emphasize the necessity of our multitask training set and prompt design, we conduct an ablation study.
We choose \texttt{BLOOMZ-7b1-mt} as the backbone model and fine-tune it using various training sets with the \textit{FixEmb-TC} method.
BLEU and COMET scores in the Zh$\Rightarrow$En direction are reported.

Our multitask training set contains three parts: \textbf{Basic Translation}, \textbf{Quality Prediction}, and \textbf{Draft Refinement}.
To demonstrate the rationality of this task combination, we remove a specific section of the training set separately, and the consequences are shown in Table \ref{tab:ablation}.
The performance of the model decreases when any subset of the training date is removed.
This result implies that each of the sub-tasks is essential for our approach.
When the Quality Prediction data is removed from the training set, the BLEU scores exhibit the most noticeable decrease.
This observation suggests that the \tree process heavily relies on the model's ability to discern various qualities of translations.

\begin{table}[t]
    \centering
    \small
    \begin{tabular}{lcccc}
        \toprule
        System & Zh$\Rightarrow$En & En$\Rightarrow$Zh & De$\Rightarrow$En & En$\Rightarrow$De \\
        \midrule
        ICL-2shot & 77.43 & 78.69 & 82.99 & 78.85 \\
        ICL-3shot & 77.89 & 79.60 & 83.05 & 79.27 \\
        ICL-4shot & 77.91 & 79.89 & 83.16 & 79.65 \\
        \tree & 79.53 & 84.24 & 84.11 & 83.80 \\
        \bottomrule
    \end{tabular}
    \caption{COMET scores gained by our approach and the ICL method.}
    \label{tab:taste_vs_icl}
\end{table}

\subsection{Comparison with Related Methods}
\paragraph{\tree vs CoT}
Our approach is based on a two-stage inference, which is similar to the thought of CoT.
To certify the superiority of our proposal, we perform a comparison with the CoT method.
We apply the same prompts utilized in \tree to guide a two-stage inference process with \texttt{LLaMA-2-chat-7b} and \texttt{LLaMA-2-chat-13b}, both of which undergo no fine-tuning process.
The results are shown in Table \ref{tab:taste_vs_cot}.
In many-to-English translation directions, the ICL method gains reasonable performance, yet our approach outperforms it significantly.
In English-to-many directions, the ICL method failed to generate stable outcomes by the inference chain, primarily due to a severe off-target issue that kept the models from producing translations in correct target languages.

\begin{table}[t]
    \centering
    \begin{tabular}{lcc}
        \toprule
        Period & De$\Rightarrow$En & En$\Rightarrow$De \\
        \midrule
        Before & 78.27 & 72.06 \\
        After & 84.16 & 84.19 \\
        \bottomrule
    \end{tabular}
    \caption{COMET scores obtained before and after the post-editing process.}
    \label{tab:ape}
\end{table}

\paragraph{\tree vs ICL}
We also conduct a comparative analysis between \tree and ICL methodologies.
We employ \texttt{LLaMA-2-chat-7b} as the backbone model and incorporate source-target pairs randomly sampled from the Base Translation training set as examples within the prompts.
The ICL experiment encompasses settings ranging from 2-shot to 4-shot scenarios.
2-shot to 4-shot settings are involved in the experiment.
The results, showcased in Table \ref{tab:taste_vs_icl}, reveal a significant performance margin between the ICL methods and our \tree approach.

\subsection{\tree as an APE Tool}

In the proposed \tree framework, the fine-tuned LLMs are employed for the evaluation and refinement of their \textbf{own} draft translations.
This naturally leads to the question: \textit{Are the fine-tuned} \tree \textit{LLMs able to evaluate base translations generated by \textbf{arbitrary systems} and refine them as an Automatic Post-Editing (APE) tool?}

To answer this question, we conducted an experiment utilizing \tree as an automatic post-editing tool.
Initially, we select \texttt{BLOOMZ-7b} in the \textit{MT-FixEmb} baseline setting to generate base translations. Subsequently, we employ \texttt{LLaMA-2-7b} in the \textit{FixEmb-TC} setting as the APE model.
We concatenate the base translation behind the prompt for the first inference stage and input it into the APE model to generate the quality label.
Finally, we format the base translation and quality label into the prompt for the second inference stage to obtain the refined translation.

The results of this experiment, as indicated by the COMET scores before and after APE, are detailed in Table \ref{tab:ape}.
Notable quality enhancements through the APE process can be observed, and the results even outperform the \tree \texttt{LLaMA-2-7b} model due to the multi-system voting mechanism.
This indicates that \tree can not only serve as an effective inference framework for a single LLM but also as an APE tool to enhance translations generated by other translation systems.

\section{Conclusion}
We introduce \tree, a novel approach that enables LLMs to translate through the self-reflection process.
Our approach allows LLMs to initially generate a preliminary translation and autonomously assess its quality.
Subsequently, the translation is refined based on the evaluation results, resulting in the final candidate.
Our experiments and analyses provide evidence of the effectiveness of \tree, as it successfully enhances the translation quality through the refinement process, consistently producing high-quality candidates across various translation directions.
Furthermore, our findings underscore that LLMs possess significant potential for the translation quality prediction task. The translation process can leverage this capacity to discern different qualities among translations, leading to the generation of high-quality outcomes.

\section*{Limitations}
The performance enhancement introduced by our approach exhibits inconsistency across different translation directions.
We assume that this phenomenon is caused by the inherent uneven multilingual knowledge within the model, and a more in-depth exploration of the underlying principles is warranted. 
Additionally, considering the two inference stages in the \tree process, the computation cost is twice that of the conventional translation generation process.
However, it's worth noting that this extra time consumption can be mitigated through acceleration methods, such as quantification and speculative decoding.

\section*{Acknowledgements}
This work was supported in part by the National Natural Science Foundation of China (Grant No. 62206076), Guangdong Basic and Applied Basic Research Foundation (Grant No. 2024A1515011491), Shenzhen Science and Technology Program (Grant Nos. ZDSYS20230626091203008 and KJZD20231023094700001). Xuebo Liu was sponsored by CCF-Tencent Rhino-Bird Open Research Fund. We would like to thank the anonymous reviewers and meta-reviewer for their insightful suggestions.

\bibliography{anthology,custom}

\appendix
\section{Appendix}

\subsection{Quality Prediction Task Designs} \label{sec:quality_prediction}

The quality prediction task is designed in two forms: text classification (TC) and quality estimation (QE).

\paragraph{Text Classification (TC)}
We instruct LLMs to categorize translations into three classes by the instruction ``\texttt{Translate from [SRC] to [TGT], and label the translation quality as ``Good'', ``Medium'' or ``Bad''.}''
For the candidates with the top 10\% COMET scores, the gold labels are assigned as ``Good'', while those with the bottom 50\% of COMET scores are labeled as ``Bad''.
Candidates falling within the remaining range are designated as ``Medium''.

\paragraph{Quality Estimation (QE)}
We request LLMs to simultaneously predict integer quality scores ranging from 0 to 100 while generating translations by the following instruction: ``\texttt{Translate from [SRC] to [TGT], and score the translation quality from 0 to 100.}'' Here, the placeholders ``\texttt{[SRC]}'' and ``\texttt{[TGT]}'' denote the source and target language, respectively.
We amplify the COMET scores by a factor of one hundred and round them to use as gold scores.

The QE task can be regarded as a more precise version of the TC task, which is perceived as more challenging for generative language models.
The methodologies employed during the training and test phase will remain consistent.

\begin{table}[t]
    \centering
    \begin{tabular}{lccc}
        \toprule
        Task & Good & Medium & Bad \\
        \midrule
        Quality Prediction & 30.0k & 30.0k & 30.0k \\
        Draft refinement & 8.0k & 8.0k & 4.0k \\
        \bottomrule
    \end{tabular}
    \caption{Numbers of instances of all three quality categories in the training set for the Quality Prediction and Draft Refinement sub-task.}
    \label{tab:label_num}
\end{table}

\begin{table}[t]
    \centering
    \begin{tabular}{lcc}
        \toprule
        Task & Size & Source \\
        \midrule
        Basic Translation & 45.4k & WMT Dev \\
        Draft Refinement & 20.0k & MTME \\
        Quality Prediction & 90.0k & MTME \\
        \bottomrule
    \end{tabular}
    \caption{Data sizes and sources of the training sets.}
    \label{tab:data_size}
\end{table}

\begin{table*}[t]
    \centering
    \scalebox{0.8}{
    \begin{tabular}{lcccccccccc}
        \toprule
        \multirow{2}{*}{\textbf{Model Size}} & \multicolumn{2}{c}{\textbf{Zh$\Rightarrow$En}} & \multicolumn{2}{c}{\textbf{En$\Rightarrow$Zh}} & \multicolumn{2}{c}{\textbf{De$\Rightarrow$En}} & \multicolumn{2}{c}{\textbf{En$\Rightarrow$De}} & \multicolumn{2}{c}{\textbf{Average}} \\
        \cmidrule(lr){2-3} \cmidrule(lr){4-5} \cmidrule(lr){6-7} \cmidrule(lr){8-9} \cmidrule(lr){10-11}
         & COMET & BLEU & COMET & BLEU & COMET & BLEU & COMET & BLEU & COMET & BLEU \\
        \midrule
        MT-\textit{FixEmb} & 79.84 & 23.43 & 85.20 & 36.68 & 78.27 & 25.07 & 72.06 & 19.41 & 78.84 & 26.15 \\
        \tree & & & & & & & & \\
        \quad\textit{FixEmb-QE} & 80.40 & 24.41 & 85.81 & 39.31 & 79.20 & 26.28 & 76.30 & 21.84 & 80.43 & 27.96 \\
        \quad\textit{FixEmb-TC} & 80.28 & 24.20 & 85.90 & 39.07 & 78.96 & 26.27 & 76.38 & 21.98 & 80.38 & 27.88 \\
        \quad\textit{FixEmb-Mix-QE} & 79.97 & 24.26 & 85.65 & 38.87 & 78.63 & 26.29 & 75.19 & 21.15 & 79.86 & 27.64 \\
        \quad\textit{FixEmb-Mix-TC} & 80.11 & 24.19 & 85.60 & 38.73 & 78.48 & 25.90 & 75.04 & 21.02 & 79.81 & 27.46 \\
        \bottomrule
    \end{tabular}}
    \caption{COMET and BLEU scores achieved by the merged model. \textit{FixEmb-Mix-QE} and \textit{FixEmb-Mix-TC} represent the results obtained by the merged model employing QE and TC approaches during the inference process, respectively.}
    \label{tab:mix_model}
\end{table*}

\begin{figure*}[t]
    \centering
    \includegraphics[width=0.85\linewidth]{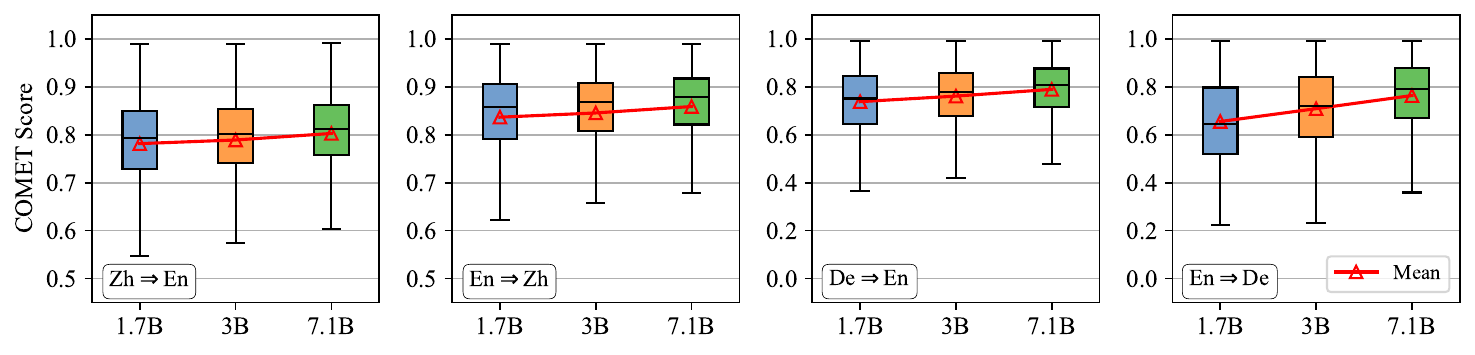}
    \caption{COMET scores obtained from \texttt{BLOOMZ} across different model sizes.}
    \label{fig:model_size}
\end{figure*}

\begin{table*}[!t]
    \centering
    \scalebox{0.8}{
    \begin{tabular}{lcccccccccc}
        \toprule
        \multirow{2}{*}{\textbf{Model Size}} & \multicolumn{2}{c}{\textbf{Zh$\Rightarrow$En}} & \multicolumn{2}{c}{\textbf{En$\Rightarrow$Zh}} & \multicolumn{2}{c}{\textbf{De$\Rightarrow$En}} & \multicolumn{2}{c}{\textbf{En$\Rightarrow$De}} & \multicolumn{2}{c}{\textbf{Average}} \\
        \cmidrule(lr){2-3} \cmidrule(lr){4-5} \cmidrule(lr){6-7} \cmidrule(lr){8-9} \cmidrule(lr){10-11}
         & COMET & BLEU & COMET & BLEU & COMET & BLEU & COMET & BLEU & COMET & BLEU \\
        \midrule
        1.7B & 78.15 & 20.76 & 83.67 & 34.96 & 73.82 & 21.80 & 65.53 & 17.21 & 75.29 & 23.68 \\
        3B & 78.91 & 22.54 & 84.56 & 36.43 & 76.14 & 23.88 & 70.90 & 19.12 & 77.63 & 25.49 \\
        7.1B & 80.28 & 24.20 & 85.90 & 39.07 & 78.96 & 26.27 & 76.38 & 21.98 & 80.38 & 27.88 \\
        \bottomrule
    \end{tabular}}
    \caption{COMET and BLEU scores achieved by \texttt{BLOOMZ} across different model sizes.}
    \label{tab:model_size}
\end{table*}

\subsection{Data Details} \label{sec:data_details}
\paragraph{WMT Development Data}
We use human-written validation data from previous WMT competitions as the basic MT training data to align LLMs on the MT task.
Specifically, we choose the newstest2017-2021 of German $\Leftrightarrow$ English and Chinese $\Leftrightarrow$ English as our MT training set.

\paragraph{MTME Multi-Candidate Data}
This is a dataset containing source sentences and translation candidates of multiple MT systems on the WMT Metrics Shared Tasks built by Google Research.
We use the candidates of newstest2019-2021 in German $\Leftrightarrow$ English and Chinese $\Leftrightarrow$ English directions to build training data for the Quality Prediction and Draft Refinement task.
For Quality Prediction, the inputs for the LLMs are the instructions and the source sentences, and the text generation labels are sampled candidates with their corresponding quality labels/scores attached at the end.
For Draft Refinement, we choose the candidate with the highest COMET score among all candidates of one source sentence as the label for the LLMs, and the draft translation is sampled from the rest of them.
The inputs for the LLMs are the instructions, the source sentences, and the drafts with their corresponding quality labels/scores attached in the front.

\begin{table*}[t]
    \centering
    \begin{tabular}{p{3cm}p{13.3cm}}
        \toprule
        \multicolumn{1}{c}{Task} & \multicolumn{1}{c}{Prompt} \\
        \midrule
        \multicolumn{1}{c}{\multirow{7}{*}{\shortstack[c]{Basic \\ Translation}}} & Write a response that appropriately completes the request.\nl\nl \#\#\#Request:\nl \\
         & Translate from Chinese to English.\nl \\
         & \begin{CJK}{UTF8}{gbsn} 一辆 1948 年的福特水星汽车穿过佐治亚州门罗小镇的一群围观者，朝着小小的摩尔滩桥隆隆奔行。\end{CJK}\nl\nl \\
         & \#\#\# Note: A translation with no errors could be\nl\nl \\
         & \#\#\# Response: A 1948 Ford Mercury passed through a group of onlookers in rural Monroe, Georgia, and rumbled toward the small Moore's Ford Bridge. \\
        \midrule
        
        \multicolumn{1}{c}{\multirow{6}{*}{\shortstack[c]{Quality\\ Prediction\\ (TC)}}} & Write a response that appropriately completes the request.\nl\nl \#\#\#Request:\nl \\
         & Translate from English to German, and label the translation quality as ``Good'', ``Medium'' or ``Bad''\nl \\
         & \begin{CJK}{UTF8}{gbsn} 北京大兴国际机场首航开启了北京“双机场”时代。\end{CJK}\nl\nl \\
         & \#\#\# Response: The first flight of Beijing Daxing International Airport ushered in the era of Beijing's ``double airport.''\nl[Good] \\
        \midrule
        
        \multicolumn{1}{c}{\multirow{5}{*}{\shortstack[c]{Quality\\ Prediction\\ (QE)}}} & Write a response that appropriately completes the request.\nl\nl \#\#\#Request:\nl \\
         & Translate from Chinese to English, and score the translation quality from 0 to 100.\nl \\
         & \begin{CJK}{UTF8}{gbsn} 7月26日在上海拍摄的公共卫生防疫专业委员会成立仪式现场。\end{CJK}\nl\nl \\
         & \#\#\# Response: The scene of the inauguration ceremony of the Public Health Epidemic Prevention Professional Committee taken in Shanghai on July 26.\nl [83] \\
        \midrule

        \multicolumn{1}{c}{\multirow{8}{*}{\shortstack[c]{Draft \\ Refinement}}} & Write a response that appropriately completes the request.\nl\nl \#\#\#Request:\nl \\
         & Translate from Chinese to English. \\
         & \begin{CJK}{UTF8}{gbsn} 虽然朱雨玲连追3分，但丁宁还是利用发球以11：9拿下首局。\end{CJK}\nl\nl \\
         & \#\#\# Hint:\nl Draft with quality label:\nl [Bad] Although he had only three points, he took the ball to 11:9. \\
         & \#\#\# Note: A translation with no errors could be\nl\nl \\
         & \#\#\# Response: Although Zhu Yuling chased three points in a row, but Ding Ning used his serve to take the first set 11-9. \\
        \bottomrule
    \end{tabular}
    \caption{Examples of the prompts and labels for the LLMs. We follow \citet{jiao2023parrot} to surround the inputs with ``Write a response that appropriately completes the request.\textbackslash n\textbackslash n\#\#\# Request:\textbackslash n'' and ``\#\#\# Response:'' to guide the LLMs to complete specific tasks. The contents behind ``\#\#\# Response:'' are the labels for the text generation fine-tuning of the LLMs.}
    \label{tab:prompt_example}
\end{table*}

To enable the LLMs to have a good understanding of the translation quality, we carefully designed the proportion of the candidates with different quality levels.
We classified the candidates into three categories by the COMET scores evaluated by \texttt{wmt-22-comet-da}.
Candidates with the top 10\% COMET scores are classified as ``Good'', while those with the bottom 50\% of COMET scores are classified as ``Bad''.
Candidates falling within the remaining range are designated as ``Medium''.
For the Quality Prediction and Draft Refinement training set, the numbers of instances constructed by candidates of all three quality categories are shown in Table \ref{tab:label_num}.

The sizes and sources of the training data for the three tasks are represented in Table \ref{tab:data_size}.
Examples of the complete prompts and labels for these tasks are shown in Table \ref{tab:prompt_example}.

\subsection{Merged Model} \label{sec:merged_model}
We also train a model that merges two types of Quality Prediction approaches, Text Classification (TC) and Quality Estimation (QE), to facilitate the \tree self-reflection process and generate both preliminary and refined translations.
Users have the flexibility to specify the approach by instructing the model in the first inference stage.
If the instruction is ``\texttt{Translate from [SRC] to [TGT], and label the translation quality as ``Good'', ``Medium'' or ``Bad''}'', then the TC approach is adopted, and the model predicts quality labels for the preliminary translation.
Otherwise, if the instruction is ``\texttt{Translate from [SRC] to [TGT], and score the translation quality from 1 to 100}'', the model employs the QE approach and predicts quality scores.
For training the merged model, we utilized 45.4k instances of Basic Translations, 45k instances for each of the two Quality Prediction approaches (TC and QE), and 20k instances for each of the two Draft Refinement styles (TC and QE).
\texttt{BLOOMZ-7b1-mt} is employed as the backbone model.

The results are shown in Table \ref{tab:mix_model}.
We observe that although there is a marginal decrease in translation performance, the merged model demonstrates the capability to handle two types of quality expression approaches simultaneously and successfully conducts the normal inference process as the non-merged models.

\subsection{Effect of Model Size} \label{sec:model_scaling}
We report the COMET and BLEU scores yielded by \texttt{BLOOMZ} of various model sizes in Figure \ref{fig:model_size} and Table \ref{tab:model_size}.

We can observe that with the increase in the number of model parameters, both the median and mean scores are consistently rising. 
This indicates that our proposed method is robust in terms of model parameter scaling.
As mentioned in \S\ref{sec:results}, LLMs depend on large amounts of parameters to memorize task-specific knowledge to perform multi-tasking.
In addition, the instructions we designed for different tasks are highly similar, which makes it more challenging but essential for LLMs to grasp different types of knowledge.

Another observation is that the distribution of scores achieved by larger models tends to be more concentrated than that obtained by smaller ones.
This indicates that as the number of model parameters increases, the performance of LLMs is not only enhanced but also stabilized, which means bad cases occur less frequently, guaranteeing the lower bound of the capacity.

\begin{table*}[t]
    \centering
    \begin{tabular}{lp{14cm}}
        \toprule
        \multicolumn{2}{c}{\textbf{Case 1}} \\
        \textbf{SRC} & \begin{CJK}{UTF8}{gbsn} 扫二维码下载贴吧客户端\end{CJK} \\
        \textbf{REF} & Scan QR code to download Tieba app \\
        \textbf{Stage 1} & Sweep the two-dimensional code to download the client \colorbox{yellow}{[Bad]} \\
        \textbf{Stage 2} & Scan the QR code to download the client \\
        \midrule
        \multicolumn{2}{c}{\textbf{Case 2}} \\
        \textbf{SRC} & \begin{CJK}{UTF8}{gbsn} “部级领导干部历史文化讲座”20周年精华珍藏版 王春瑜、张大可、熊召政、葛剑雄、邓小南等 14位史学泰斗、文化名家、知名学者献给部级领导干部的公开课\end{CJK} \\
        \textbf{REF} & “History and Culture Lectures for Ministerial Cadres” 20th Anniversary Essence Collector’s Edition, an open class dedicated to ministerial cadres by 14 leading historians, cultural masters and well-known scholars including Wang Chunyu, Zhang Dake, Xiong Zhaozheng, Ge Jianxiong, Deng Xiaonan \\
        \hdashline
        \textbf{Stage 1} & "20th Anniversary of the History and Culture Lecture of Leading Cadres at the Ministerial Level" \colorbox{yellow}{[Bad]} \\
        \textbf{Stage 2} & The 20th Anniversary of the "Historical and Cultural Lectures for Leading Cadres at the Ministerial Level" - 14 Famous Historians, Cultural Celebrities, and Renowned Scholars Including Wang Chunyu, Zhang Dake, Xiong Zhengxiong, and Deng Xiaonan Present Public Lectures for Leading Cadres at the Ministerial Level \\
        \midrule
        \multicolumn{2}{c}{\textbf{Case 3}} \\
        \textbf{SRC} & \begin{CJK}{UTF8}{gbsn} 当时刚创立的台积电，在英特尔这个巨无霸面前就是小喽罗。\end{CJK} \\
        \textbf{REF} & The newly established Taiwan Semiconductor Manufacturing Company was only a small player in front of Intel the giant. \\
        \hdashline
        \textbf{Stage 1} & At that time, the newly established Taiwan accumulator, in the face of the giant Intel, is a small squeak. \colorbox{yellow}{[Bad]} \\
        \textbf{Stage 2} & At that time, the newly established Taiwan Semiconductor was just a small fry in the face of Intel, the giant. \\
        \midrule
        \multicolumn{2}{c}{\textbf{Case 4}} \\
        \textbf{SRC} & \begin{CJK}{UTF8}{gbsn} 比如专业、年龄、想学习的语言，甚至在哪个城市，这些都会和你学习或者想要从事的行业有关系的。\end{CJK} \\
        \textbf{REF} & Such as your major, age, the language you want to learn, and even which city you are in; all of these are related to what you want to learn or what industry you want to go into. \\
        \hdashline
        \textbf{Stage 1} & For example, professional, age, want to learn the language, even in which city, these will be related to your learning or want to engage in the industry. \colorbox{yellow}{[Medium]} \\
        \textbf{Stage 2} & For example, your major, your age, the language you want to learn, and even the city you want to study in, are all related to the industry you want to work in. \\
        \midrule
        \multicolumn{2}{c}{\textbf{Case 5}} \\
        \textbf{SRC} & \begin{CJK}{UTF8}{gbsn} 箱中袋装水，其包装盒采用100％可回收利用，可减少包装66％，减少碳排放97％。\end{CJK} \\
        \textbf{REF} & The packaging of the bagged water in the box is 100\% recyclable, which can reduce packaging by 66\% and carbon emission by 97\%. \\
        \hdashline
        \textbf{Stage 1} & The packaging box uses 100\% recyclable packaging, which can reduce packaging by 66\% and reduce carbon dioxide emissions by 97\%. \colorbox{yellow}{[Good]} \\
        \textbf{Stage 2} & The bagged water in the box is packed in a 100\% recyclable packaging box, which can reduce packaging by 66\% and carbon dioxide emissions by 97\%. \\
        \bottomrule
    \end{tabular}
    \caption{Cases of translation process of \tree in Chinese $\Rightarrow$ English direction. The backbone model is \texttt{LLaMA-2-7b} trained with its embedding layer fixed. ``Stage 1'' represents the preliminary translation generated during the first inference process, and ``Stage 2'' represents the refined translation generated during the second inference process. The predicted quality labels for the drafts are marked using \colorbox{yellow}{highlights}.}
    \label{tab:case_study}
\end{table*}

\subsection{Case Study}
Several cases of the translation process of \tree are shown in Table \ref{tab:case_study}.
The quality labels are predicted as ``Bad'' while some severe lexical or syntactic problems occur in the preliminary translations.
In Case 1, an ambiguous Chinese character ``\begin{CJK}{UTF8}{gbsn}扫\end{CJK}'' is inaccurately translated into ``sweep'', and the term ``\begin{CJK}{UTF8}{gbsn}二维码\end{CJK}'' is literally translated as ``two-dimensional code'' instead of ``QR code''.
In Case 2, the preliminary translation is incomplete, omitting the latter part of the source sentence.
In Case 3, the word order in the preliminary translation is notably awkward.
All these issues are effectively addressed during the second inference stage, resulting in refined translations of high quality.
Even when the predicted quality labels are designated as ``Medium'' or ``Good'', as seen in Case 4 and Case 5, the second stage inference continues to perform fine-tuning on the preliminary translations based on the actual context and linguistic nuances.

\end{document}